\documentclass[11pt]{article}

\usepackage[final]{acl}

\usepackage{times}
\usepackage{latexsym}
\usepackage{hyperref}
\usepackage{amsmath, amsthm}
\usepackage{booktabs}

\usepackage{enumitem}
\usepackage{amsmath, amsthm}
\usepackage{algorithm}
\usepackage{algpseudocode}
\usepackage{amssymb}
\usepackage{tabularx}
\newcolumntype{Y}{>{\centering\arraybackslash}X}

\usepackage[T1]{fontenc}
\usepackage[utf8]{inputenc}

\usepackage{microtype}

\usepackage{inconsolata}

\usepackage{graphicx}

\theoremstyle{definition} 
\newtheorem{definition}{Definition}[section] % [section] resets numbering at each section

\title{Towards Reliable, Generalizable, and Specific In-Context Knowledge Editing via Multi-Objective Reinforcement Learning}

\author{
  Xuzhong Wang$^1$,
  Maiqi Jiang$^1$,
  Tejal Nair$^1$,
  Girija Bhusal$^2$,
  Yanfu Zhang$^1$,
  \textbf{Haipeng Chen}$^1$ \\
  $^1$College of William and Mary \quad $^2$Tribhuvan University \\
  \texttt{\{xwang58, mjiang04, tnair, yzhang105, hchen23\}@wm.edu} \\
  \texttt{girija.bhusal9@gmail.com}
}

\begin{document}
\maketitle
\begingroup
\renewcommand\thefootnote{}\footnote{Code is publicly available at \url{https://github.com/wmd3i/MO-IKE}.}%
\addtocounter{footnote}{-1}%
\endgroup
\begin{abstract}
Large Language Models (LLMs) are powerful but limited by static parametric knowledge that becomes outdated once pretraining ends. Knowledge editing addresses this problem by updating model behavior on target facts without full retraining. In particular, \textbf{in-context knowledge editing} has gained attention because it is training-free and readily applicable to black-box LLMs. Recent reinforcement learning (RL)-based approaches improve over fixed retrieval strategies by adapting prompt construction to the quantity--quality trade-off. Despite initial success, they optimize \textbf{reliability} alone, and in doing so trade away \textbf{specificity}: rewarding edit success drives the retriever to discard the demonstrations that protect neighboring facts. Previous methods also act over only part of the prompt construction process, overlooking the global organization of demonstrations. We propose \textbf{M}ulti-\textbf{O}bjective \textbf{I}n-context \textbf{K}nowledge \textbf{E}diting (\textbf{MO-IKE}), which casts prompt construction as a sequential decision process over COPY, UPDATE, RETAIN, and STOP actions, and trains a dynamic retriever with a multi-objective shaped reward that couples edit success to explicit paraphrase and retention penalties. On Llama-3.2-3B, averaged over five seeds, MO-IKE raises edit success (reliability) from 87.1\% to 91.1\% and retention rate (specificity) from 41.0\% to 63.4\% over the strongest RL baseline, while keeping paraphrase consistency (generality) comparable (79.1\% to 77.7\%), improving the harmonic-mean score from 61.7 to 75.7. Gains hold across five frozen LLMs and four datasets, including the 311K-example UniEdit benchmark.
\end{abstract}

\section{Introduction}
Large Language Models (LLMs) have become prevalent in natural language processing, excelling in domains ranging from autonomous workflows \cite{schick2023toolformer} to advanced mathematical reasoning \cite{chervonyi2025goldmedalistperformancesolvingolympiad}. However, a critical limitation remains: parametric knowledge in LLMs is static. Once training concludes, the stored information becomes susceptible to obsolescence \cite{mazzia2024surveyknowledgeeditingneural}. For instance, an LLM whose training precedes the 2026 Winter Olympics lacks parametric knowledge of the final medal tally. Consequently, relying solely on static memory invariably fails to satisfy user needs for real-time or dynamic information. 

\textit{Knowledge editing} addresses this by updating an LLM’s response to a target fact while preserving unrelated knowledge \cite{wang2024knowledgeeditinglargelanguage}. While gradient-based methods \cite{mitchell2022fast} achieve this via weight updates, they are computationally expensive and cannot be adapted to black-box LLMs \cite{meng2022locating}. Recently, \textit{in-context knowledge editing} has become popular due to its advantage of being training-free and naturally applicable to black-box LLMs \cite{zheng2023ike}: the model remains frozen, and the desired update is induced solely through prompt context, without access to model internals. 

The efficacy of in-context knowledge editing is highly dependent on the injected prompt. Thus, a critical question is \emph{how to construct the prompt}. Precisely, the context must be reliable enough to successfully edit the LLM (\textbf{reliability}), general enough to apply to semantically equivalent queries (\textbf{generality}), and specific enough to avoid changing unrelated neighboring facts (\textbf{specificity}) \cite{meng2023massediting}. Although static retrieval strategies such as IKE \cite{zheng2023ike} propose constructing the prompt with different categories of demonstrations—specifically, COPY demonstrations that explicitly state the new fact for reliability, UPDATE demonstrations that state a paraphrased query for generality, and RETAIN demonstrations that state an unrelated fact for specificity—these methods remain insufficient, as they assume a fixed amount of context is adequate for distinct edits. 

\begin{figure}
    \centering
    \includegraphics[width=\linewidth]{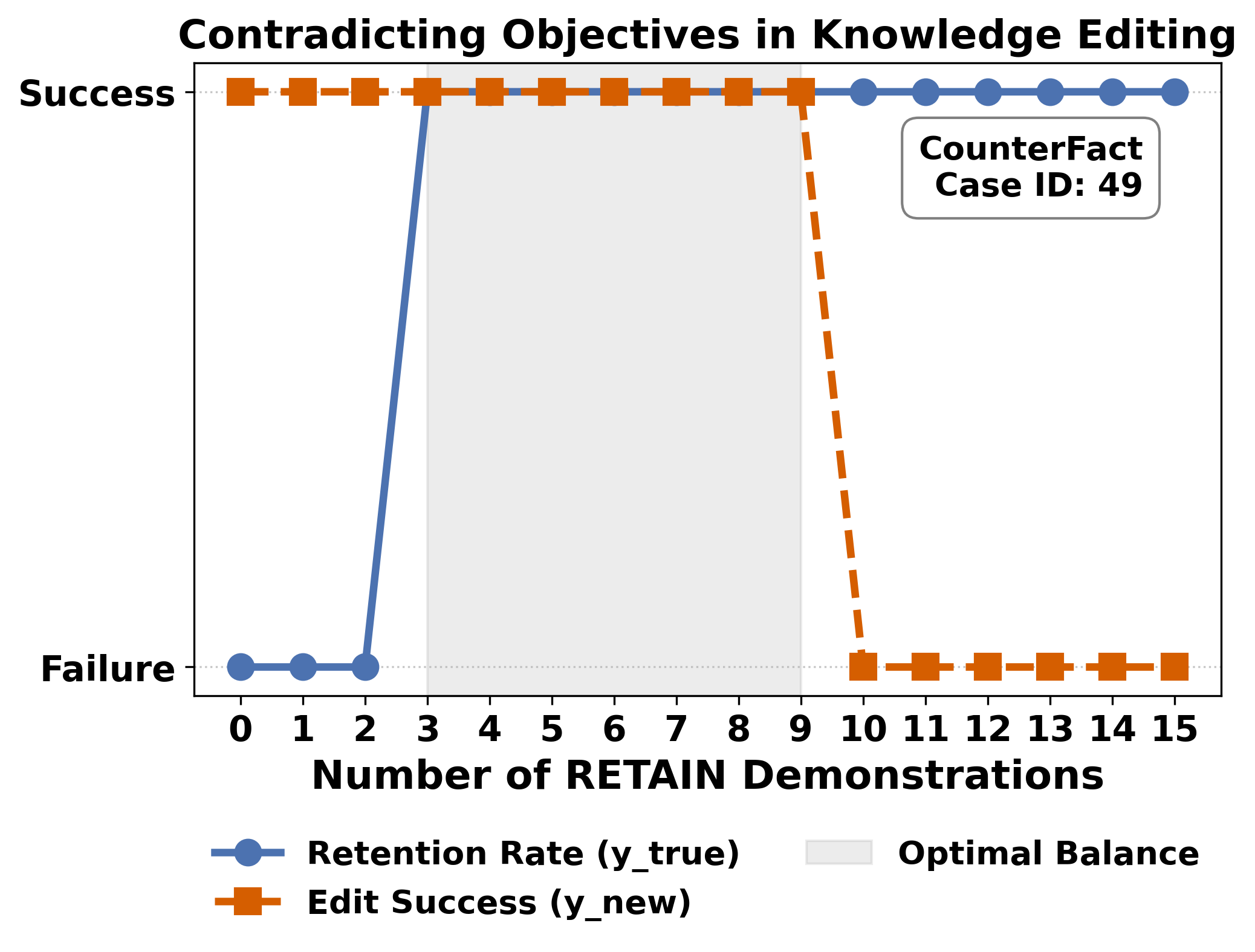}
    \caption{Contradicting objectives of reliability and specificity in prompt construction in IKE -- Increasing the number of RETAIN demonstrations leads to Retention Success but Edit Failure.}
    \label{fig: trade-off}
\end{figure}

To address this limitation, recent work such as Dynamic Retriever for In-Context Knowledge Editing (DR-IKE) \cite{nafee-etal-2025-dynamic} formulates demonstration selection as a reinforcement learning (RL) problem, enabling the system to adapt the number of retrieved examples based on each edit. Despite their initial success, DR-IKE exhibits two limitations. First, while it improves \emph{how much} context is supplied, it leaves partially explored \emph{how that context should be organized} by ranking only RETAIN candidates, as the effectiveness of in-context learning is known to be sensitive to demonstration ordering \cite{lu-etal-2022-fantastically}.
% as auto-regressive models exhibit primacy and recency biases over prompt position \cite{wang2024learningretrieveincontextexamples}. 

Second, DR-IKE does not account for the \textit{proportion} of demonstrations from distinct categories. Effective in-context prompt construction requires a \textit{principled balance} among distinct demonstration types to achieve reliability, generality, and specificity. From the observation in Figure~\ref{fig: trade-off}, these objectives are often contradictory \cite{zheng2023ike}. On the one hand, achieving strong reliability and generality demands sufficient COPY and UPDATE examples to reinforce the factual change \cite{qiao-etal-2024-comem}. On the other hand, maintaining specificity requires RETAIN examples that anchor unedited neighboring knowledge \cite{youssef-etal-2025-make}. Under a limited context budget, we have to carefully balance the proportion of different demonstrations and consider potential trade-offs. 

To alleviate the dilemma of competing objectives, we formulate the retrieval environment as a \textbf{Constrained Markov Decision Process (Constrained MDP)} and introduce \textbf{MO-IKE} as its solution algorithm. We formulate the state as the currently constructed prompt, and the action of the retriever is to either select an additional demonstration or terminates the prompt construction. We evaluate the final prompt based on its edit reliability given the constraints on generality and specificity. During training, MO-IKE samples a group of candidate prompts under the current policy and optimizes the retriever via Group Relative Policy Optimization (GRPO). 

Our main contributions are as follows:
\begin{itemize}[noitemsep,topsep=0pt,leftmargin=*]
    \item We identify two notable deficiencies in state-of-the-art RL-based in-context knowledge editing (notably, DR-IKE): they neglect the sequential ordering of the prompt context and overlook the balance of examples from different demonstration categories. 
    
    \item We propose \textbf{MO-IKE}, a Multi-Objective RL algorithm that jointly optimizes the selection of distinctive demonstration categories. By framing the task as a Constrained MDP and optimizing it with multi-objective RL, we ensure stable training that balances reliability, generality, and specificity.
    
    \item We introduce architectural improvements to a dynamic demonstration retriever. On standard knowledge-editing benchmarks, including \textsc{CounterFact}, using Llama-3.2-3B and Mistral-7B-v0.3, MO-IKE improves edit success (reliability) by up to +7.0\% (reaching 92.0\% on Llama-3.2), paraphrase consistency (generality) by up to +2.5\%, while yielding a +23.0\% absolute improvement in retention rate (specificity).
\end{itemize}

\section{Related Work}\label{section: related}

\subsection{In-Context Knowledge Editing}
Early work on knowledge editing largely relies on gradient-based approaches. For instance, MEND \cite{mitchell2022fast} trains a smaller editor network for targeted weight updates. Constrained by the heavy computational overhead \cite{wang2024wise}, researchers explore in-context knowledge editing as a gradient-free alternative. Initial explorations improve prompt engineering via prefixing \cite{cohen-etal-2024-evaluating}, chain-of-thought \cite{wang-etal-2025-knowledge-editing}, and formalized IKE framework \cite{zheng2023ike}, which utilizes k-NN to retrieve diverse \textsc{COPY}, \textsc{UPDATE}, and \textsc{RETAIN} demonstrations to address reliability, generality and specificity, respectively. 
% Subsequent efforts have sought to make this framework more robust. For example, IKE has been enhanced with rewriting agents for unstructured tasks \cite{ziboxu2025eikeaenhancing}, shifted toward decoding-time interventions rather than prompt modification \cite{bi-etal-2024-adaptive}, and augmented with continuous persuasion tokens to avoid lengthy context requirements \cite{youssef2026persuasiontokenseditingfactual}. 
Most relevantly, DR-IKE \cite{nafee-etal-2025-dynamic} utilizes RL to train a BERT-based retriever \cite{devlin2019bertpretrainingdeepbidirectional} to dynamically construct a prompt. Yet, while these approaches successfully improve edit reliability, they rarely address the balance between reliability, generality, and specificity.
\vspace{1mm}

A natural question is whether IKE is distinct from Retrieval-Augmented Generation (RAG). The two settings have underlying objectives that differ: RAG aims to \emph{supplement} the model's parametric knowledge with external documents \cite{lewis2021retrievalaugmentedgenerationknowledgeintensivenlp,asai2023selfrag}. IKE, by contrast, aims to \emph{overwrite} specific parametric facts while preserving unrelated knowledge. The survey of
\citet{wang2024knowledgeeditinglargelanguage} draws this boundary explicitly, concluding that RAG is unsuitable for the targeted, fact-level updates that knowledge editing seeks to achieve. As a consequence, RL-based RAG retrievers \citep{Huang_2026} are not designed to navigate the reliability--specificity--retention tradeoff that defines the IKE problem. We further discuss the question in Appendix~\ref{appendix: ragrl}.

\subsection{RL for Demonstration Selection}
Selecting optimal demonstrations for in-context learning (ICL) is challenging due to the combinatorial nature of the search space \cite{purohit2025sample} and the prompt's sensitivity to example interactions \cite{gupta-etal-2023-coverage}, demonstration order \cite{lu-etal-2022-fantastically}, and model biases \cite{xiang-etal-2024-addressing}. Because static heuristics like embedding similarity \cite{liu-etal-2022-makes} fail to capture interdependency, recent work has reframed inference-time optimization as a sequential decision-making process. RL algorithms—such as PPO \cite{schulman2017proximalpolicyoptimizationalgorithms} and GRPO \cite{shao2024deepseekmathpushinglimitsmathematical}—have driven good advances in this direction. Beyond improving reasoning \cite{xu2025largereasoningmodelssurvey, liu2025bag}, RL has proven highly effective for enhancing tool use \cite{qian2025toolrl}, training web agents \cite{qi2025webrl}, optimizing query rewriting \cite{ma-etal-2023-query}, managing agent memory \cite{yan2026memoryr1enhancinglargelanguage}, and accelerating speculative decoding \cite{wang2026speculativesamplingreinforcementlearning}.

However, applying RL to balance the competing objectives of knowledge editing remains underexplored. DR-IKE \cite{nafee-etal-2025-dynamic} pioneered this transition by formulating the IKE demonstration selection as an RL process. Yet, by optimizing primarily for a single objective (reliability), DR-IKE leads the retriever to suffer degradation in specificity and knowledge retention. While Constrained MDPs \cite{ganguly2025provablyefficientsamplecomplexity} and multi-objective optimization frameworks \cite{efroni2025alignedmultiobjectiveoptimization, efroni2025aligned} theoretically address such trade-offs, their practical application to inference-time selection is limited. 

\section{Problem Statement}
\begin{definition}\label{def: knowledge editing}
(Knowledge Editing).
Fix a frozen language model $\mathcal{M}$ and a factual triple $\mathcal{K_C}$ encoded in the model's parametric memory. Knowledge editing $f$ pursues a post-edited model $\mathcal{M'} := f(\mathcal{M}, \mathcal{K_C}\to \mathcal{K'_C})$ such that:
\begin{itemize}[noitemsep,topsep=0pt,parsep=0pt,partopsep=0pt,leftmargin=*]
    \item \textbf{Reliability.} For the exact query $q$ targeting the original fact $\mathcal{K}_C$, the output of $\mathcal{M}'$ accurately reflects the revised fact $\mathcal{K}'_C$.
    \item \textbf{Generality.} For any query $q^*$ that is semantically equivalent to $q$ (e.g., paraphrases) or whose answer logically depends on $\mathcal{K}_C$, the output of $\mathcal{M}'$ is consistent with $\mathcal{K}'_C$.
    \item \textbf{Specificity.} For any query $q'$ that depends on unrelated knowledge $\mathcal{K}_S$ (i.e., $\mathcal{K}_C \cap \mathcal{K}_S = \emptyset$), the output of $\mathcal{M}'$ aligns identically with the unedited model $\mathcal{M}$.
\end{itemize}
\end{definition}

\begin{definition} (\textbf{In-context Knowledge Editing (IKE)}). Fix a language model $\mathcal{M}$ with frozen parameters. In-context knowledge editing utilizes the base prompt $\mathcal{P}$ and a retriever to concatenate an augmented Prompt $P^*$
\begin{equation}
\mathcal{P^*} := \mathcal{P} + <d_1, d_2, \ldots, d_n>  
\end{equation}
for $d_i$ from $d_1$ to $d_n$ that are natural language demonstrations selected from the example pool by the retriever.
\end{definition}
To satisfy the competing constraints of reliability, generality, and specificity, IKE formulates the construction of the prompt $P^*$ as follows

\begin{definition} (Demonstration Categories \cite{zheng2023ike})
Each retrieved demonstration $d_i$ from $d_1$ to $d_n$ is assigned to one of the three categories.

\begin{itemize}[noitemsep,topsep=0pt,parsep=0pt,partopsep=0pt,leftmargin=*]
    \item \textbf{COPY} - directly restates the targeted fact. 

    Example: \textit{The newest iPhone} is $\rightarrow$ \textbf{iPhone 17}.
    \item \textbf{UPDATE} - paraphrases the query before stating the targeted fact.

    Example: \textit{The latest iPhone in the generation is} $\rightarrow$ \textbf{iPhone 17}.
    \item \textbf{RETAIN} - states the neighboring fact that should not change.

    Example: \textit{The newest iPad is} $\rightarrow$ \textbf{iPad M5 Pro}.
\end{itemize}
\end{definition}

\section{Methodology}

Our primary goal is to train a retriever that can construct contexts to effectively overwrite LLM's stored parametric knowledge. Ideally, we would desire the edited LLM's output to be the newly injected fact for any query targeted at that specific information. Moreover, a successful edit should prevent the neighboring facts from being modified. Nevertheless, we identify that prior baselines, notably DR-IKE, only considers editing signal in the RL training. An obvious drawback of the previous approach is that editing specificity signal is totally neglected. Thus, the resulting policy is susceptible to ``reward hacking'' where the trained retriever filtered all RETAIN candidates in favor for editing success. To achieve a principled balance among different objectives, we formulate the training process as a \textit{Constrained Markov Decision Process (Constrained MDP)}: besides edit reliability, we have to consider the constructed prompt's impact on paraphrase consistency and knowledge retention. To operationalize this idea, we employ cost constraints in addition to editing reward to measure the retriever's performance across all objectives. The final scalar reward signal is computed via multi-objective RL through individual rewards of reliability, generality, and specificity. Throughout the training process, our multi-objective framework helps the retriever optimize for a stable policy that balances distinctive and often competing objectives for an effective knowledge editing. We formalize this procedure as MO-IKE, a multi-objective in-context knowledge editing algorithm.

\subsection{IKE as a Constrained MDP}
To guarantee editing generality and specificity, the retriever should be aware of these constraints in the training. Hence, we model the sequential demonstration selection process as a Constrained MDP $(\mathcal{S}, \mathcal{A}, \mathcal{T}, \mathcal{R}, \mathcal{C})$. The retriever operates sequentially over discrete time steps $t$, constructing a prompt to query the frozen LLM.

\noindent\textbf{State.} The state $s_t \in \mathcal{S}$ represents the prompt context at step $t$. It is constructed by appending a newly selected demonstration $d_t$, formatted into a natural language string via a mapping function $\phi(\cdot)$, to the previous state. Thus, $s_0$ is the original query $q$, and $s_t = s_{t-1} \oplus \phi(d_t)$, where $\oplus$ denotes string concatenation.

\noindent\textbf{Action.} In contrast to DR-IKE, which considers only RETAIN candidates in its action space \cite{nafee-etal-2025-dynamic}, we formulate the action space over the full set of COPY, UPDATE, and RETAIN demonstrations. We introduce this change in order to encourage the exploration of a ``global'' optimal ordering between distinctive demonstration types. At step $t$, an action $a_t$ selects a demonstration without replacement from the available pool $\mathcal{D}_t$, or chooses a learnable pseudo-token $a_{stop}$ to halt. The action space is thus $\mathcal{A}_t = \mathcal{D}_t \cup \{a_{stop}\}$, where $\mathcal{D}_{t+1} = \mathcal{D}_t \setminus \{a_t\}$.

\noindent\textbf{State Transition.} The transition function $\mathcal{T}(s_{t+1} | s_t, a_t)$ is deterministic. If the agent selects a demonstration $a_t \in \mathcal{D}$, it is appended to the current state, yielding $s_{t+1} = s_t \oplus a_t$. If the agent selects $a_t = a_{\text{stop}}$, the episode immediately terminates, and the constructed prompt is finalized for evaluation.

\begin{figure*}
    \centering
    \includegraphics[width=\linewidth]{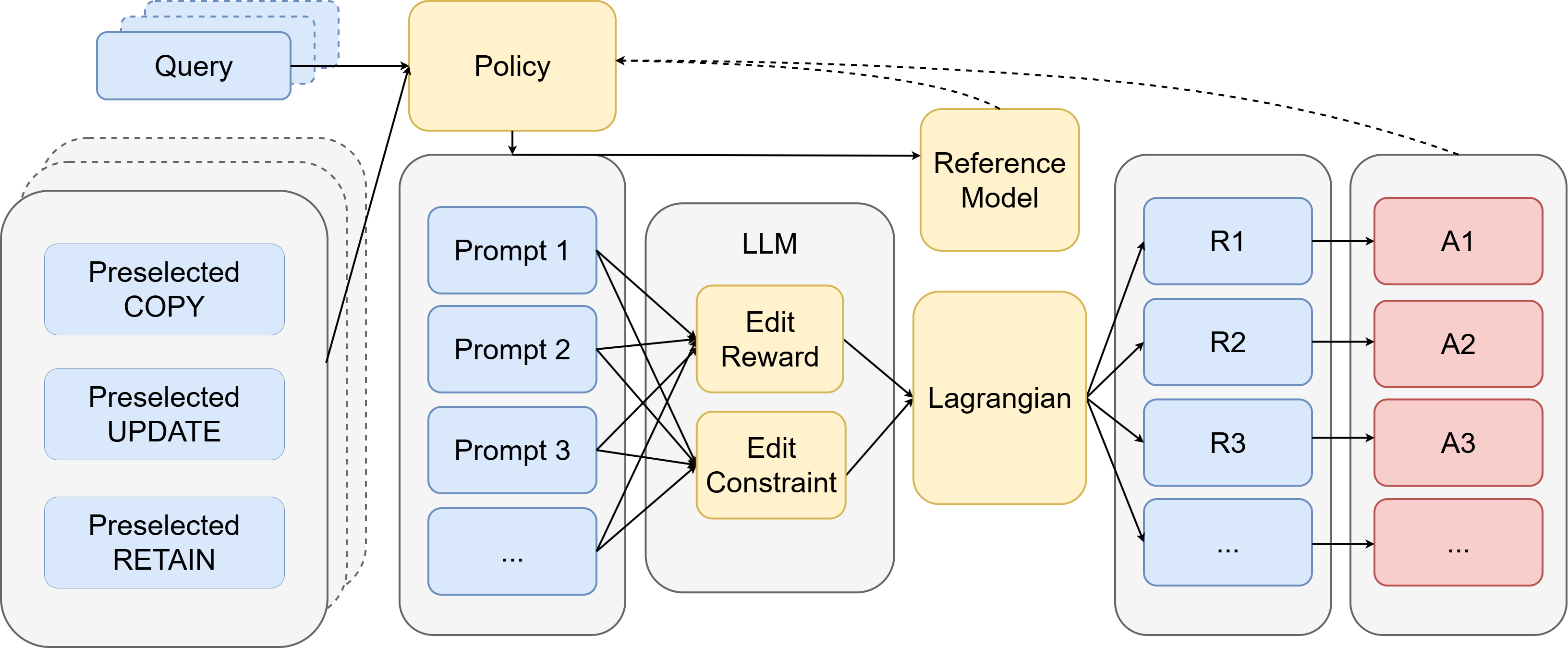}
    \caption{\textbf{MO-IKE overview.} Given a query, we first preselect COPY, UPDATE, and RETAIN candidates using kNN, then sample prompts under the current retriever policy. The LLM evaluates these prompts to produce rewards and constraint penalties, which are combined via Lagrangian relaxation into final rewards $R_1, R_2, R_3, \dots$. These rewards are then used to compute group advantages $A_1, A_2, A_3, \dots$ for policy optimization. Dashed lines mark operations used for training, not inference.}    
    \label{fig: training pipeline}
\end{figure*}

\noindent\textbf{Reward.} Our primary objective is Edit Success (ES), which represents edit reliability. Specifically, the primary reward is defined as
\begin{equation}
    R(s_t, a_t) = \mathbf{1}[\hat{y}^{edit} = y_{new}],
\end{equation}
where $\hat{y}^{edit}$ denotes the frozen LLM's prediction on the edit instance by the constructed prompt, and $y_{new}$ denotes the desired edited target.

\noindent\textbf{Cost Constraints.} Different from prior baselines, we introduce cost constraints so as to ``regularize'' RL training. We treat Paraphrase Consistency (PC), which represents generality, and Retention Rate (RR), which represents specificity as constraints for our retriever. 
Let $\hat{y}^{para}$ and $\hat{y}^{retain}$ denote the frozen LLM's predictions on the paraphrase query and retention query, respectively. Let $y_{new}$ denote the desired edited target and $y_{true}$ denote LLM's original stored answer for neighboring facts. 
We define each cost as the logical inverse of task success:
\begin{equation}
    C_{PC}(s_t, a_t) = 1 - \mathbf{1}[\hat{y}^{para} = y_{new}],
\end{equation}
\begin{equation}
    C_{RR}(s_t, a_t) = 1 - \mathbf{1}[\hat{y}^{retain} = y_{true}].
\end{equation}

\subsection{Multi-Objective Reward Shaping}

To obtain the final objective, we scalarize the multi-objective problem into a single shaped reward, using a fixed-multiplier Lagrangian relaxation as the bridge. We define a composite reward that penalizes degradation on PC and RR:
\begin{equation}
r(s_t, a_t) = R(s_t, a_t) - \sum_{k \in \{\text{PC}, \text{RR}\}} \lambda_k C_k(s_t, a_t),
\end{equation}
where $\lambda_{\text{PC}}$ and $\lambda_{\text{RR}}$ are fixed hyperparameters rather than dual variables updated by ascent. We therefore do not solve a constrained program; the relaxation serves to \emph{derive} the reward structure, and the resulting objective is multi-objective reward shaping optimized directly with GRPO. We give the derivation in Appendix~\ref{sec: appendix lagrangian}.

\subsection{MO-IKE}

In general, MO-IKE constructs editing prompts through a four-stage pipeline: \textit{candidate retrieval}, \textit{sequential selection}, \textit{reward evaluation}, and \textit{policy optimization}. (i) First, for a given edit query, we retrieve an initial candidate pool of factual demonstrations. (ii) Next, a BERT-based \cite{devlin2019bertpretrainingdeepbidirectional} retriever sequentially selects samples from this pool to build the prompt step-by-step, halting only when it chooses a ``stop'' token. (iii) The fully constructed prompt is then evaluated against both editing success (reliability), paraphrase consistency (generality) constraints, and retention (specificity). (iv) These competing objectives are combined using a fixed-penalty coefficient, producing a final scalar reward. We illustrate the core idea of MO-IKE in Algorithm~\ref{algo:mo_ike_summary}, and attach the full detailed Algorithm in Appendix~\ref{sec: appendix algo}.

\noindent\textbf{Demonstration Selection.} For a given edit instance $(x, y_{\text{new}})$, we construct the prompt via a two-stage retrieval pipeline. First, we use a Sentence Transformer \cite{reimers-gurevych-2019-sentence} to extract an initial pool of \textsc{Copy}, \textsc{Update}, and \textsc{Retain} candidates via kNN. Our BERT-based retriever then performs fine-grained sequential selection from this pool to build the final prompt.

\begin{algorithm}[t]
  \caption{MO-IKE}
  \label{algo:mo_ike_summary}
  \begin{algorithmic}[1]
    \Require Retriever $\pi_\theta$, dataset $\mathcal{D}_{\text{train}}$, group size $G$, fixed penalty $\lambda$
    \For{each $(x, y_{\text{new}}) \in \mathcal{D}_{\text{train}}$}
    
      % \State 
      \Comment{\textbf{\textit{Sampling \& Reward Evaluation:}}}
      \State Sample $G$ prompt trajectories $\{\tau_g\}_{g=1}^G \sim \pi_\theta(\cdot \mid x)$ based on current policy $\pi$
      \State Compute composite reward $r_g$ for each $\tau_g$ using fixed penalty $\lambda$
      
      \Comment{\textbf{\textit{Advantage Estimation:}}}
      \State Compute group-normalized advantages: $A_g \leftarrow (r_g - \mu) / \sigma$
      
      \Comment{\textbf{\textit{Policy Update:}}}
      \State Update $\theta$ via gradient descent to minimize:
      \Statex \hspace{\algorithmicindent}$\mathcal{J}(\theta) = \frac{1}{G} \sum_{g=1}^{G} \big[ \mathcal{L}_{\text{clip}}(\theta, \tau_g, A_g)$
      \Statex \hspace{\algorithmicindent}\hspace{3.2em}$- \beta_{\mathrm{KL}}\mathbb{D}_{\mathrm{KL}}(\pi_\theta \,\|\, \pi_{\theta_{\text{old}}}) \big]$
      \EndFor
  \end{algorithmic}
\end{algorithm}

\noindent\textbf{Policy and Action Sampling.} The MO-IKE retriever uses a frozen 4-layer BERT encoder \cite{devlin2019bertpretrainingdeepbidirectional} with trainable projection matrices for the query ($W_p$) and demonstrations ($W_d$). We introduce a learned termination pseudo-embedding $v_{\text{stop}}$ to represent the STOP action $a_{\text{stop}}$. As a sequential decision making process, the policy samples based on current state $p_t$ as a partially constructed prompt, and the candidates pool $\mathcal{A}_t$. At step $t$, with a partially constructed prompt $p_t$ from the previous $t-1$ steps and candidate pool $\mathcal{A}_t$, the selection probability for any action $a_i \in \bar{\mathcal{A}}_t$ is computed via a softmax over projected dot products:
\begin{equation}
    P(a_i | p_t) = \frac{\exp(v_p^\top v_i)}{\sum_{a_j \in \bar{\mathcal{A}}_t} \exp(v_p^\top v_j)}
\end{equation}
where $v_p = W_p \cdot \text{BERT}(p_t)$ and $v_i = W_d\cdot \text{BERT}(a_i)$. During training, we sample $a \sim P(\cdot | p_t)$ to encourage exploration. At inference, we select the highest-probability demonstration. If $a_{\text{stop}}$ is chosen, prompt construction terminates. Otherwise, we update the state ($p_{t+1} = p_t \oplus a$) and shrink the pool ($\mathcal{A}_{t+1} = \mathcal{A}_t \setminus \{a\}$).

\noindent\textbf{Policy Update.} For each instance $(x, y_{\text{new}})$, we sample a group of $G$ prompt trajectories $\{ \tau_1, \dots, \tau_G \}$ from the current policy $\pi_\theta$. We evaluate these trajectories using the target LLM to compute a composite reward $r_i$ based on edit success, paraphrase consistency, and retention metrics. Group-relative advantages are estimated as $A_i = (r_i - \mu_r) / \sigma_r$, where $\mu_r$ and $\sigma_r$ are the group's reward mean and standard deviation. We optimize $\theta$ by maximizing the clipped surrogate objective, penalized by the Kullback-Leibler (KL) divergence from a reference policy $\pi_{\text{ref}}$ (the initial, reference retriever weights):
\begin{equation}
\mathcal{L}_{\text{GRPO}}(\theta) = \mathcal{L}_{\text{clip}}(\theta) - \beta \mathbb{D}_{\text{KL}}[\pi_\theta \| \pi_{\text{ref}}],
\end{equation} where $\mathcal{L}_{\text{clip}}(\theta)$ is given by:
\begin{equation}
    \begin{aligned}
\mathcal{L}_{\text{clip}}(\theta) = \frac{1}{G} \sum_{i=1}^G \min \Bigg( \frac{\pi_\theta(\tau_i | q)}{\pi_{\text{old}}(\tau_i | q)} A_i, \\
\text{clip} \Bigg( \frac{\pi_\theta(\tau_i | q)}{\pi_{\text{old}}(\tau_i | q)}, 1-\epsilon, 1+\epsilon \Bigg) A_i \Bigg)
\end{aligned}
\end{equation}

\section{Experiments}
\label{sec: experiments}

\subsection{Experimental Setup}
\noindent\textbf{Datasets.} In this section, we present evaluations on MO-IKE and our baselines across three standard knowledge editing benchmarks: \textsc{CounterFact} \cite{meng2022locating}, \textsc{ZsRE} \citep{levy-etal-2017-zero}, and \textsc{Wiki-Counterfact} \citep{cohen-etal-2024-evaluating}. For each benchmark, we partition the factual records into two distinct sets: an editable sample pool for evaluation and a separate pool for constructing IKE demonstrations. Following \citet{zheng2023ike}, we allocate the first 2,000 samples (out of 21,919) from \textsc{CounterFact} to the editable pool. We apply an analogous split to the other datasets, reserving the first 500 samples from both \textsc{ZsRE} (out of 11,230) and \textsc{WikiData\textsubscript{counterfact}} (out of 2,340) for editing. The remaining records in each dataset are utilized to build the demonstration pools. We additionally evaluate MO-IKE on larger datasets UniEdit \cite{chen2025unieditunifiedknowledgeediting}, which is shown more in details in Appendix~\ref{appendix: add_experiments}.

\noindent\textbf{Language Models.} We evaluate our approach using several instruction-tuned large language models: Meta-Llama-3.1-8B-Instruct, Meta-Llama-3.2-3B-Instruct \cite{grattafiori2024llama3herdmodels}, Mistral-7B-Instruct-v0.2 \cite{jiang2023mistral7b}, and Qwen2.5 (1.5B and 7B) \cite{yang2024qwen2technicalreport}. All model parameters remain strictly frozen throughout our experiments.

\noindent\textbf{Training Configuration.} Similar to \citet{nafee-etal-2025-dynamic}, we train our retriever on 300 randomly selected examples and evaluate it on held-out sets of 300 samples for \textsc{CounterFact}, and 100 samples each for \textsc{ZsRE} and \textsc{WikiData\textsubscript{counterfact}}. We provide detailed training configuration in Appendix~\ref{appendix: configuration}.

\noindent\textbf{Baselines.} We compare our approach against four representative in-context knowledge editing baselines: \textbf{FactPrompt} \citep{cohen-etal-2024-evaluating}, which prepends a narrative prefix to guide editing; \textbf{EditCoT} \citep{wang-etal-2025-knowledge-editing}, which leverages chain-of-thought prompting; \textbf{IKE} \citep{zheng2023ike}, which categorizes demonstrations into COPY, UPDATE, and RETAIN types; and \textbf{DR-IKE} \citep{nafee-etal-2025-dynamic}, which utilizes policy optimization for dynamic demonstration selection.

\noindent\textbf{Evaluation Metrics.} We evaluate MO-IKE across three dimensions of knowledge editing: reliability, generality, and specificity. Let $y_{new}$ and $y_{true}$ denote the newly injected fact and the original stored truth, respectively.

\begin{table*}[t]
\small
\centering
\begin{tabularx}{\textwidth}{|l|Y|Y|Y|Y|Y|Y|Y|}
\hline
\textbf{Editing Method} & \textbf{S ↑} & \textbf{ES ↑} & \textbf{EM ↑} & \textbf{PC ↑} & \textbf{PM ↑} &  \textbf{RR ↑} & \textbf{RM ↑}  \\
\hline
\multicolumn{8}{|l|}{\textbf{Llama-3.2-3B-Instruct}} \\ 
\hline
FactPrompt & 34.8 & 55.3 & 64.0 & 25.7 & 23.8 & 34.3 & -7.1 \\ \hline
EditCoT & 42.5 & 70.5 & 62.4 & 43.5 & 21.6 & 29.9 & 1.0 \\ \hline
IKE & \underline{63.9} & 78.0 & 67.9 & 68.0 & 60.0 & 52.0 & \underline{29.9} \\ \hline
DR-IKE & 54.9 & \underline{85.0} & \underline{76.4} & 77.0  & \textbf{69.2} & 34.7 & -4.2 \\ \hline
MO-IKE & \textbf{73.5} & \textbf{92.0} & \textbf{78.4} & \textbf{79.0} & \underline{66.0}& \textbf{57.7} & \textbf{30.0} \\ \hline
\multicolumn{8}{|l|}{\textbf{Mistral-7B-Instruct-v0.3}} \\ 
\hline
FactPrompt & 41.8 & 36.3 & 73.4 & 42.3 & 34.3 & 48.3 & -2.9 \\ \hline
EditCoT & 35.6 & 33.8 & 58.7 & 43.4 & 33.4 & 31.0 & 5.6 \\ \hline
IKE & \underline{64.8} & 58.0 & 84.5 & 81.3 & 69.9 & \underline{60.0} & \textbf{33.9} \\ \hline
DR-IKE & 62.0 & \underline{74.7} & \underline{91.8} & \underline{88.3} & \textbf{78.4} & 42.3 & -22.5 \\ \hline
MO-IKE & \textbf{77.4} & \textbf{80.1} & \textbf{92.0} & \textbf{90.7} & \underline{74.9} & \textbf{65.0} & \underline{32.1} \\ \hline
\end{tabularx}
 \caption{Main results of in-context knowledge editing across Llama-3.2-3B-Instruct and Mistral-7B-Instruct-v0.3. We report Overall Score (S), Edit Success (ES), Edit Metric (EM), Paraphrase Consistency (PC), Paraphrase Metric (PM), Retention Rate (RR), and Retention Metric (RM). \textbf{Bold} indicates the best performance, and \underline{underline} indicates the second best. }
\label{tab:main results}
\end{table*}

\begin{itemize}[noitemsep,topsep=0pt,leftmargin=*]
    \item \textbf{Reliability:} Measures post-editing efficacy on target prompts via Edit Success (ES), defined as $\mathbb{E}[\mathbb{I}[P(y_{new}) > P(y_{true})]]$, and Edit Magnitude (EM), defined as $\mathbb{E}[P(y_{new}) - P(y_{true})]$.
    \item \textbf{Generality:} Evaluates accuracy on paraphrased prompts via Paraphrase Consistency (PC) and Paraphrase Magnitude (PM), which are calculated identically to ES and EM, respectively.
    \item \textbf{Specificity:} Assesses the preservation of unedited neighboring facts via Retention Rate (RR), defined as $\mathbb{E}[\mathbb{I}[P(y_{true}) > P(y_{new})]]$, and Retention Magnitude (RM), defined as $\mathbb{E}[P(y_{true}) - P(y_{new})]$.
\end{itemize}

Finally, we report the overall \textbf{Score (S)} as the harmonic mean of ES, PC, and RR.

\subsection{Main results}

Table~\ref{tab:main results} presents the performance of MO-IKE on the \textsc{CounterFact} dataset. We highlight results for Llama-3.2 and Mistral-v0.3, with extended model evaluations provided in Table~\ref{tab:llm_performance}. We also include results from different datasets in Table~\ref{tab: datasets}. Furthermore, we include some additional experiments in Appendix~\ref{appendix: add_experiments}.

Across both architectures, MO-IKE consistently outperforms across the majority of metrics:
% . The core advantages of our approach are demonstrated as follows:
(1) Compared to the standard IKE baseline, MO-IKE yields substantial improvements in edit success and consistency. Against DR-IKE, it maintains or exceeds peak reliability, pushing Llama-3.2 ES from 85.0 to 92.0.
(2) RL baselines like DR-IKE often drops below the IKE baseline in RR and overall Score. MO-IKE protects existing parametric knowledge by enforcing retention. MO-IKE reverses the specificity degradation seen in prior methods. It improves RR by absolute margins of +23.0 for Llama-3.2 and +22.7 for Mistral-v0.3 over DR-IKE, while shifting Mistral's negative RM from -22.5 to a positive +32.1.
(3) As shown in Table~\ref{tab: datasets}, MO-IKE's performs well also in \textsc{ZsRE} and \textsc{Wiki-Counterfact}. It achieves the highest overall Score (42.0 on \textsc{ZsRE} and 59.0 on Wiki) while simultaneously achieving the strongest RR.

% \setlength{\tabcolsep}{6pt}
% \begin{table}[ht]
%   \centering
%   \begin{tabular}{|l|c|c|}
%     \hline
%     \textbf{Method}      & \textbf{S ↑} & \textbf{RR ↑} \\ 
%     \hline
%     \multicolumn{3}{|l|}{\textbf{ZsRE}} \\
%     \hline
%     IKE                  & \underline{34.0} & \underline{19.0}           \\ 
%     \hline
%     DR-IKE               & 33.0 & 16.0          \\ 
%     \hline
%     MO-IKE               & \textbf{42.0} & \textbf{22.0}          \\
%     \hline
%     \multicolumn{3}{|l|}{\textbf{Wiki\textsubscript{Counterfact}}} \\
%     \hline
%     IKE                  & \underline{55.0} & \underline{46.0}        \\ 
%     \hline
%     DR-IKE               & 53.0 & 33.0         \\ 
%     \hline
%     MO-IKE               & \textbf{59.0} & \textbf{47.0}         \\ 
%     \hline
%   \end{tabular}
%   \caption{Overall efficacy and knowledge preservation across the ZsRE and Wiki\textsubscript{Counterfact} datasets.}
%   \label{tab: datasets}
% \end{table}

\setlength{\tabcolsep}{6pt}
\begin{table}[ht]
  \centering
  \small
  \begin{tabular}{|l|c|c|c|c|}
    \hline
    \textbf{Method}      & \textbf{S ↑} & \textbf{ES ↑} & \textbf{PC ↑} & \textbf{RR ↑} \\ 
    \hline
    \multicolumn{5}{|l|}{\textbf{ZsRE}} \\
    \hline
    IKE                  & \underline{34.0} & 64.0           & 50.0          & \underline{19.0}           \\ 
    \hline
    DR-IKE               & 33.0 & \textbf{77.0}            & \underline{65.0}          & 16.0          \\ 
    \hline
    MO-IKE               & \textbf{42.0} & \underline{75.0}            & \textbf{79.0}          & \textbf{22.0}          \\
    \hline
    \multicolumn{5}{|l|}{\textbf{Wiki\textsubscript{Counterfact}}} \\
    \hline
    IKE                  & \underline{55.0} & 70.0 & 54.0 & \underline{46.0}        \\ 
    \hline
    DR-IKE               & 53.0 & \textbf{86.0}          & \textbf{68.0}          & 33.0         \\ 
    \hline
    MO-IKE               & \textbf{59.0} & \underline{80.0}          & \underline{59.0}         & \textbf{47.0}         \\ 
    \hline
  \end{tabular}
  \caption{Editing performance across the ZsRE and Wiki\textsubscript{Counterfact} datasets.}
  \label{tab: datasets}
\end{table}

\noindent\textbf{Zero-Shot Cross Model Evaluation.} We observe that retriever trained via MO-IKE on one language model can effectively generalize to a completely different language model without any retraining. To prove this, we training the retriever purely on Llama-3.2, and evaluating it zero-shot on Mistral-7B. As demonstrated in Table~\ref{tab:zeroshot}, MO-IKE suffers no performance drop when transferred zero-shot to Mistral-7B. In fact, it performs practically identically to the natively trained version, outperforming all other baselines. 
\begin{table}[ht]
 \small
  \centering
    \resizebox{\columnwidth}{!}{%
  \begin{tabular}{|l|c|c|c|c|}
    \hline
    \textbf{Method}      & \textbf{S ↑} & \textbf{ES ↑} & \textbf{PC ↑} & \textbf{RR ↑} \\ 
    \hline
    IKE                  & 64.9 & 58.0 & 81.3 & 60.0 \\ 
    \hline
    DR-IKE               & 62.3 & 74.7 & 88.3 & 42.3 \\ 
    \hline
    MO-IKE (Native)      & \underline{77.1} & \underline{80.1} & \textbf{90.7} & \underline{65.0} \\ 
    \hline
    MO-IKE (Zero-Shot)   & \textbf{77.3} & \textbf{81.0} & \underline{89.3} & \textbf{65.7} \\ 
    \hline
  \end{tabular}
  }
  \caption{Zero-shot transfer performance: the retriever was trained exclusively on Llama-3.2 and evaluated zero-shot on Mistral-7B.}
  \label{tab:zeroshot}
\end{table}

\subsection{Reward Constrained Ablation}

We conduct an ablation study on the reward function to evaluate four configurations: (i) no constraints, (ii) paraphrase constraint only, (iii) retention constraint only, and (iv) all constraints. As shown in Table~\ref{tab: constraint ablation}, omitting the retention penalty (row i and ii) leads to a noticeable drop in RR. This confirms that explicitly penalizing retention degradation is necessary to prevent editing success from negatively impacting the model's pre-existing knowledge. Furthermore, the results implies a positive correlation between edit success and paraphrase consistency: optimizing for editing reliability inherently benefits generality.

\begin{table}[ht]
  \small
  \centering
  \resizebox{\columnwidth}{!}{%
  \begin{tabular}{|l|c|c|c|c|}
    \hline
    \textbf{Reward Constraint}       & \textbf{S↑} & \textbf{ES↑} & \textbf{PC↑} & \textbf{RR↑} \\ 
    \hline
    No Constraint                    &65.5  & 91.3 & 77.0 & 50.0  \\ 
    \hline
    Only Paraphrase                  &68.8  & \textbf{92.0} & \textbf{79.0} & 52.0 \\ 
    \hline
    Only Retention                   & 70.1 &86.0  &71.0  & \textbf{57.7}  \\ 
    \hline
    All Constraints                  & \textbf{73.5} & \textbf{92.0} & \textbf{79.0} & \textbf{57.7} \\ 
    \hline
  \end{tabular}%
  }
  \caption{Ablation study on the components of the MO-IKE scalarized reward function using Llama-3.2. \textbf{Bold} indicates the best performance.}
  \label{tab: constraint ablation}
\end{table}

%%% 74

It is worth noting that the full multi-objective formulation (all constraints) achieves better Score compared to only retention constraint (row iii and iv). We hypothesize that it may comes from the mechanics of group relative rewards. When the reward signal lacks multiple constraints, the group rewards tend to exhibit lower variance within a group. Consequently, the group-relative advantages become uninformative, which may downgrade the effect of the policy gradient updates. By incorporating all constraints, the reward landscape becomes more nuanced, providing the variance in advantage signals necessary for RL training. 

In addition, we discuss reward weights sensitivity in Appendix~\ref{appendix: hyperparam}.

\begin{figure}
    \centering
    \includegraphics[width=\linewidth]{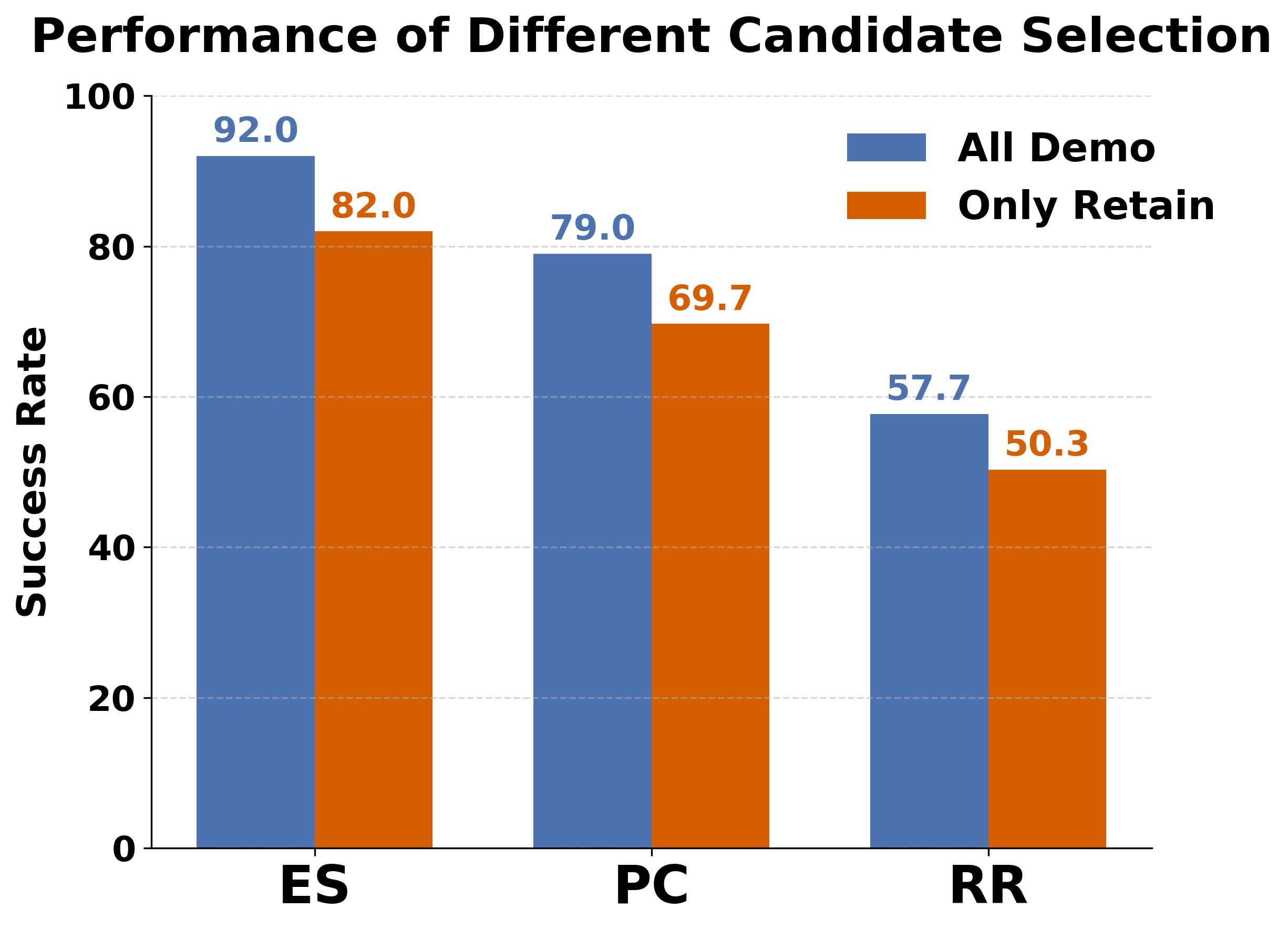}
    \caption{Performance of Llama-3.2 using different demonstration selection scopes.}    
    \label{fig: candidate selection}
\end{figure}

\subsection{Action Space Ablation}
Previous approaches to in-context knowledge editing have largely treated demonstration organization as a secondary concern. IKE \cite{zheng2023ike} constructs prompts using static format, essentially treating demonstrations as a fixed template. While DR-IKE \cite{nafee-etal-2025-dynamic} introduces a dynamic retrieval mechanism, it restricts its optimization to the ranking and construction of only RETAIN candidates, ignoring the inter-dependencies between different demonstration types. In contrast, MO-IKE's action space is expanded to include all demonstrations across all categories. By conducting RL training across the entire sequence of demonstration types—COPY, UPDATE, and RETAIN—our method captures the global, synergistic effects of context construction. As illustrated in Figure~\ref{fig: candidate selection}, our approach achieves better performance across all three metrics compared to optimizing solely for RETAIN. These empirical results confirms our decision to expand the action space across all demonstration categories.

\begin{figure}
    \centering
    \includegraphics[width=\linewidth]{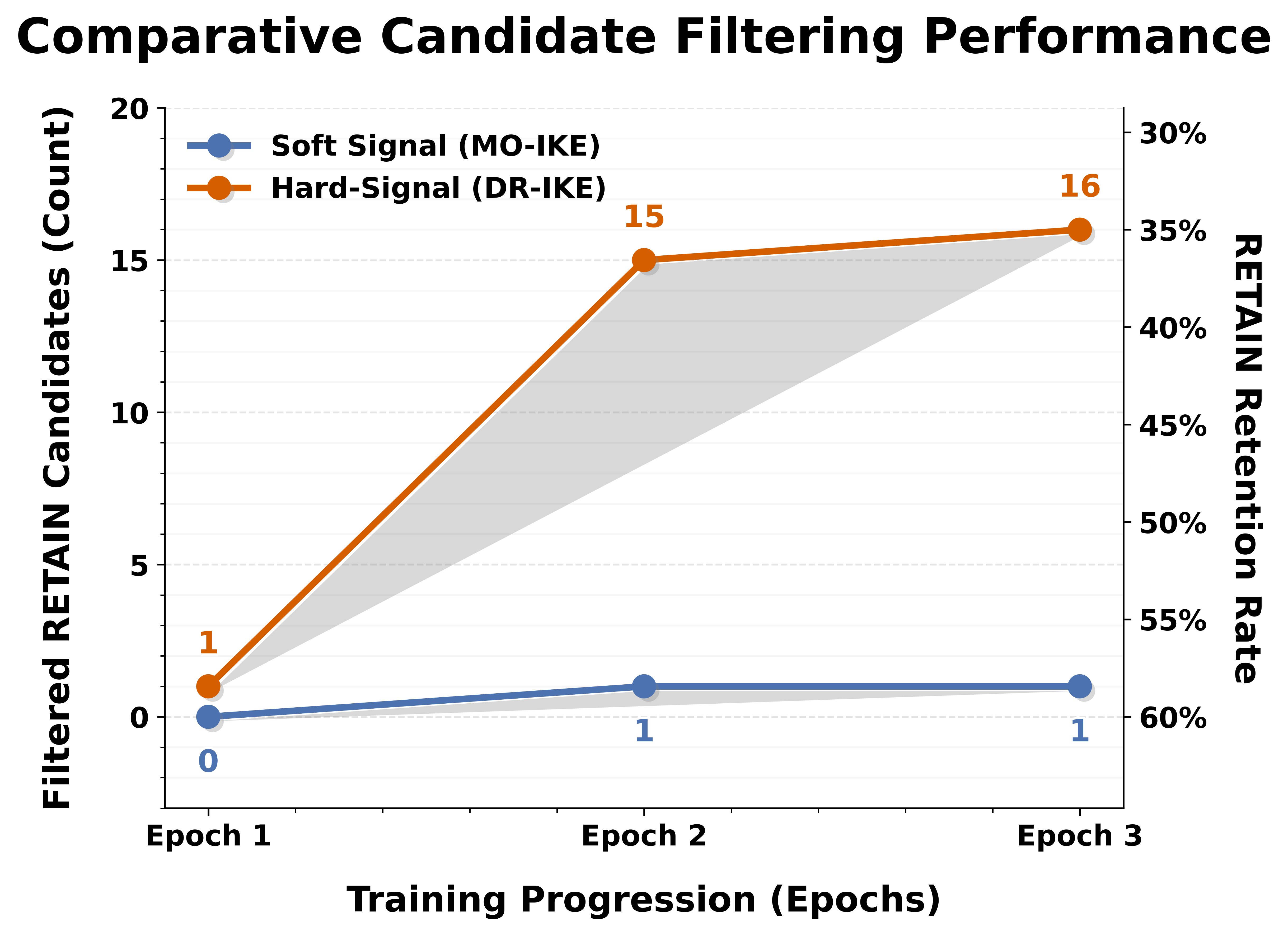}
    \caption{Left axis tracks filtered candidates; the right axis shows inverted retention degradation. The shaded gap highlights DR-IKE's hard constraint failing (over-filtering by Epoch 3) versus MO-IKE's soft embedding preserving stable retention.}    
    \label{fig: filter candidates}
\end{figure}

\subsection{Demonstration Structure and  Analysis}

The performance of our approach can be attributed to the resulting structural proportions of the demonstration candidates. DR-IKE \cite{nafee-etal-2025-dynamic} force a greedy optimization for edit success, which skews the prompt distribution by discarding RETAIN candidates. To resolve this structural imbalance, MO-IKE integrates a multi-objective RL framework and a soft stopping mechanism through a stop signal in the action space that is evaluated alongside other demonstration candidates. As shown in Figure~\ref{fig: filter candidates}, this allows the model to maintain contextual heterogeneity rather than aggressively filtering out non-edit demonstrations. Consequently, MO-IKE achieves better editing specificity overall (Table~\ref{tab:main results}). We provide detailed Case Study of the context structure in Appendix~\ref{appendix: case study}. By maintaining a balanced proportion of COPY, UPDATE, and RETAIN types, our method achieves robust results without sacrificing the overall integrity of the model's existing knowledge.

\section*{Conclusion}
We introduce \textbf{MO-IKE}, a multi-objective RL algorithm for in-context knowledge editing that trains a retriever to dynamically construct prompts. We formulate demonstration retrieval as a Constrained MDP, casting prompt construction as a sequential decision-making problem. MO-IKE explicitly optimizes the retriever over the often competing objectives of reliability, generalization, and specificity. Experiments across multiple LLMs show that MO-IKE consistently outperforms existing retrieval strategies, achieving strong effectiveness across all three objectives and cross-model generalization. Ablation studies further demonstrate the importance of our constraint formulation, as well as the embedding and ranking designs, to MO-IKE’s overall performance.

\section*{Limitations}
Several limitations still exist in our work. First, the dataset scale for knowledge editing remains relatively small; for example, we train and evaluate MO-IKE only on the first 2,000 records of the \textsc{CounterFact} dataset, which may not be sufficient to examine the model's overall capabilities. Second, by formulating the problem as a Constrained MDP, we incorporate all constraints directly into the reward function. Because we utilize fixed coefficients (e.g., a static lambda value) for these constraints, the precise impact of these hyperparameters on the RL optimization process remains unexplored. Third, although MO-IKE achieves a better retention rate overall, its absolute performance is bottlenecked by the inherent limits of in-context learning. The retention rate is still relatively low (around 60\%) compared to other metrics, suggesting this gradient-free method still has room for improvement. Future work may consider different approaches to surpass this limitation. Finally, we evaluate MO-IKE solely on a specific knowledge editing task, leaving it unclear whether this framework generalizes to other in-context learning applications, such as model reasoning. 

\section*{Acknowledgements}
This research was supported by the National Science Foundation under IIS-2348405, IIS-2451436, Commonwealth Cyber Initiative grant HC-2Q26-032, and the William \& Mary Semester Research Grant.

% Bibliography entries for the entire Anthology, followed by custom entries
%\bibliography{anthology,custom}
% Custom bibliography entries only
\bibliography{custom}

\appendix

\onecolumn

\section{Training Procedure for Multi-objective GRPO}
Algorithm \ref{algo: grpo} outlines the detailed MO-IKE algorithm.
\label{sec: appendix algo}
\begin{algorithm*}[ht]
  \caption{Multi-Objective In-Context Knowledge Editing (MO-IKE)}
  \label{algo: grpo}
  \small
  \begin{algorithmic}[1]
    \Require initial retriever parameters $\theta$, training set $\mathcal{D}_{\text{train}}$, group size $G$, clipping $\epsilon$, KL weight $\beta_{\mathrm{KL}}$
    
    \For{epoch $=1$ to $N_{\mathrm{epochs}}$}
      \For{each $(x, y_{\text{new}}) \in \mathcal{D}_{\text{train}}$}
        \State Initialize storage lists: $\mathcal{D} \leftarrow []$, $\text{rewards} \leftarrow []$
        
        \State 
        \For{$g = 1$ to $G$}
          \State $x_{\text{curr}} \leftarrow x$
          \State Sequence $R_g \leftarrow []$; \quad LogProbs $\pi^{\text{old}}_g \leftarrow []$

          \For{step $j = 1$ to $K$} \Comment{Sequentially select demonstrations}
             \State $z \leftarrow S_{\theta}(x_{\text{curr}}, C)$; \quad $p \leftarrow \text{softmax}(z)$
             \State Sample action $a_j \sim p$
             \State Append $a_j$ to $R_g$; \quad Append $p[a_j]$ to $\pi^{\text{old}}_g$
             \State Remove selected action $a_t$ from candidate pool $\mathcal{C}_t$: $\mathcal{C}_{t+1} \leftarrow \mathcal{C}_t \setminus \{a_t\}$
             \State $x_{\text{curr}} \leftarrow \text{Concat}(x_{\text{curr}}, a_j)$ \Comment{Update prompt state}
          \EndFor
          \State Prompt LLM to get $\hat{y}_g$ and reward $r_g$ \Comment{Compute composite reward}
          \State Append $r_g$ to $\text{rewards}$
          \State Store trajectory data $(R_g, \pi^{\text{old}}_g)$ in $\mathcal{D}$
        \EndFor

        \State
        \State $\mu_{\text{group}} \leftarrow \frac{1}{G}\sum r_g$ \Comment{Compute group baseline}
        \State $\sigma_{\text{group}} \leftarrow \sqrt{\frac{1}{G}\sum (r_g - \mu_{\text{group}})^2} + \delta$
        \State $A_g \leftarrow (r_g - \mu_{\text{group}}) / \sigma_{\text{group}}$ for each group $g$

        \State 
        \For{inner epoch $=1$ to $N'_{\mathrm{epochs}}$}
            \State $\mathcal{L}_{\text{total}} \leftarrow 0$
            \For{$g = 1$ to $G$}
              \State Reconstruct states $x_{g,0}, \dots, x_{g,K}$ using $x$ and $R_g$
              \For{step $j$ in $R_g$}
                \State $p^{\text{new}}_{g,j} \leftarrow \text{softmax}(S_{\theta}(x_{g,j-1}, C))_{\text{action}_j}$ 
                \State $p^{\text{old}}_{g,j} \leftarrow \pi^{\text{old}}_g[j]$ 
                \State $ratio \leftarrow p^{\text{new}}_{g,j} / p^{\text{old}}_{g,j}$     
                \State $L^{\text{CLIP}} \leftarrow -\min(ratio \cdot A_g, \text{clip}(ratio, 1-\epsilon, 1+\epsilon) \cdot A_g)$ \Comment{Compute clipped surrogate objective}
                \State $L^{\text{KL}} \leftarrow \beta_{\mathrm{KL}} \cdot \log(p^{\text{new}}_{g,j} / p^{\text{old}}_{g,j})$ \Comment{Approximate KL penalty}
                \State $\mathcal{L}_{\text{total}} \mathrel{+}= (L^{\text{CLIP}} + L^{\text{KL}})$
              \EndFor
            \EndFor
            \State $\theta \leftarrow \theta - \alpha \nabla_{\theta} (\frac{1}{G \cdot K} \mathcal{L}_{\text{total}})$ \Comment{Gradient descent update}
        \EndFor
      \EndFor
    \EndFor
  \end{algorithmic}
\end{algorithm*}

\section{Proof of the Final Reward via Lagrangian Relaxation}
\label{sec: appendix lagrangian}

We model the problem as Constrained MDP. Our goal is to find a policy $\pi_\theta$ that maximizes the Edit Success (ES) while ensuring that the expected costs associated with violating Paraphrase Consistency (PC) and Retention Rate (RR) remain below acceptable thresholds.

\subsection{Problem Formulation}
Let $J_{R}(\pi)$ denote the expected return for the primary editing task based on the reward $R(s_t, a_t)$. We introduce cost constraints for PC and RR, where $J_{C_k}(\pi)$ represents the expected cumulative cost for constraint $k \in \{\text{PC}, \text{RR}\}$. Let $d_k$ represent the maximum tolerable expected cost for that metric.

The CMDP optimization problem is defined as:
\begin{equation}
\begin{aligned}
\max_{\pi_\theta} \quad & J_{R}(\pi_\theta) \\
\text{s.t.} \quad & J_{C_k}(\pi_\theta) \le d_k, \quad \forall k \in \{\text{PC}, \text{RR}\}
\end{aligned}
\end{equation}
where the expectations are taken over the trajectory distribution induced by the policy $\pi_\theta$.

\subsection{Lagrangian Relaxation}
 We apply the method of Lagrange multipliers. We define the \textbf{Lagrangian objective function} $\mathcal{L}(\pi, \lambda)$ as:
\begin{equation}
\mathcal{L}(\pi, \lambda) = J_{R}(\pi) - \sum_{k \in \{\text{PC}, \text{RR}\}} \lambda_k (J_{C_k}(\pi) - d_k)
\end{equation}
Here, $\lambda = (\lambda_{\text{PC}}, \lambda_{\text{RR}})$ is a vector of Lagrange multipliers, with $\lambda_k \ge 0$. The primal constrained problem is equivalent to the following unconstrained \textbf{min-max} dual problem:
\begin{equation}
\min_{\lambda \ge 0} \max_{\pi_\theta} \mathcal{L}(\pi, \lambda)
\end{equation}

\subsection{Equivalence to Scalarized Reward}
We can rewrite the Lagrangian function by expanding the expected return and expected cost terms over the trajectories $\tau \sim \pi$. Since expectations are linear, we have:
\begin{equation}
\begin{aligned}
\mathcal{L}(\pi, \lambda) &= \mathbb{E}{\tau \sim \pi} \left[ \sum{t} R(s_t, a_t) \right] - \sum_{k \in {\text{PC}, \text{RR}}} \lambda_k \left( \mathbb{E}{\tau \sim \pi} \left[ \sum{t} C_{k}(s_t, a_t) \right] - d_k \right) \\
&= \mathbb{E}{\tau \sim \pi} \left[ \sum{t} \left( R(s_t, a_t) - \sum_{k \in {\text{PC}, \text{RR}}} \lambda_k C_{k}(s_t, a_t) \right) \right] + \sum_{k \in {\text{PC}, \text{RR}}} \lambda_k d_k
\end{aligned}
\end{equation}
Notice that the term $\sum_{k} \lambda_k d_k$ is constant with respect to the policy optimization step (the inner maximization loop over $\pi$). Therefore, maximizing $\mathcal{L}(\pi, \lambda)$ with respect to $\pi$ is mathematically equivalent to maximizing the expected return of a new, \textbf{composite reward function}:
\begin{equation}
r(s_t, a_t) = R(s_t, a_t) - \sum_{k \in \{\text{PC}, \text{RR}\}} \lambda_k C_k(s_t, a_t)
\end{equation}
Thus, by optimizing this composite reward using GRPO with fixed hyperparameters $\lambda_{\text{PC}}$ and $\lambda_{\text{RR}}$, we are equivalently solving the inner loop of the Lagrangian dual problem. The fixed coefficients $\lambda_k$ act as the Lagrange multipliers that apply static penalty weightings to balance the trade-off between editing performance and the cost constraints.
\twocolumn

\section{Comparison to RL-Based RAG Retrieval}
\label{appendix: ragrl}

To test whether RL-based RAG retrievers can substitute for an IKE-specific approach, we evaluate RAG-RL \cite{Huang_2026} on CounterFact with Qwen-2.5-7B-Instruct. RAG-RL was originally trained to optimize answer-F1 and citation-F1 on open-domain QA; we apply it directly to the IKE setting without modification. We observe that the LLM frequently produces refusals (``I cannot determine the answer based on the prompt'') when RAG-RL's retrieval forces it into contradiction with its parametric knowledge.

\setlength{\tabcolsep}{6pt}
\begin{table}[ht]
  \centering
  \begin{tabular}{|l|c|c|c|c|}
    \hline
    Methods      & \textbf{S ↑} & \textbf{ES ↑} & \textbf{PC ↑} & \textbf{RR ↑} \\ 
    \hline
    RAG-RL               & 20.0 & 83.0 & 79.3 & 8.0 \\ 
    \hline
    MO-IKE               & \textbf{78.1} & \textbf{96.0} & \textbf{88.3} & \textbf{60.0} \\ 
    \hline
  \end{tabular}

  \caption{RAG-RL evaluated on CounterFact with Qwen-2.5-7B-Instruct. RAG-RL achieves moderate edit success but collapses on retention (8.0\% RR), consistent with our claim in Section~\ref{section: related}. }
  \label{tab:ragrl}
\end{table}

\section{Training Configuration}
\label{appendix: configuration}
To ensure reproducibility, Table \ref{tab:hyperparameters} details the complete set of hyperparameters used during the training of MO-IKE. 

\begin{table}[h]
\centering
\begin{tabular}{lc}
\toprule
\textbf{Hyperparameter} & \textbf{Value} \\
\midrule
Learning Rate & $1\times 10^{-5}$ \\
Epochs & 3 \\
Group Size & 8 \\
Random Seed & 42 \\
GRPO Clipping Parameter ($\epsilon$) & 0.1\\
KL Coefficient ($\beta$) & 0.001 \\
Lagrangian Penalty Weights & 1 \\
\bottomrule
\end{tabular}
\caption{Hyperparameters used for MO-IKE training.}
\label{tab:hyperparameters}
\end{table}

\section{Training Dynamics}

\begin{figure}
    \centering
    \includegraphics[width=\linewidth]{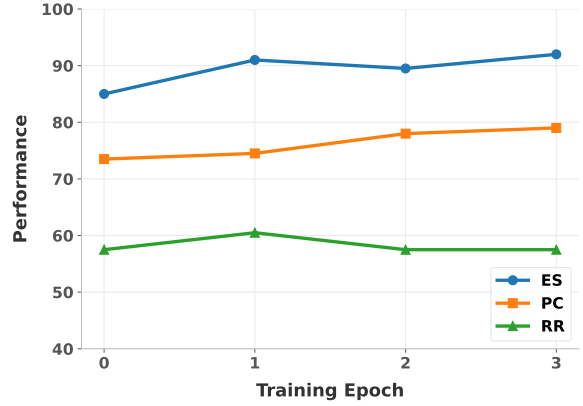}
    \caption{Training dynamics with Llama-3.2. We evaluate the performance of demonstration selection given different training epochs.}
    \label{fig: convergence}
\end{figure}

To understand the learning efficiency of our RL-based retrieval optimization, we analyze the performance metrics of MO-IKE across training epochs. Figure~\ref{fig: convergence} illustrates the progression of ES, PC and RR over the duration of the training process. 

A key observation is the rapid convergence of the policy. By the conclusion of the first epoch, the model establishes a strong, balanced demonstration ordering, evidenced by the sharp initial peak in both ES and RR. Throughout subsequent epochs (Epoch 2 and 3), the performance largely plateaus; ES and PC demonstrate stable refinements, while RR maintains its rate without degrading. Because the reward includes a penalty for retention, policy updates are regularized toward smaller deviations, which can reduce unstable exploratory shifts during early RL training \cite{schulman2017trustregionpolicyoptimization}. The model rapidly identifies a near-optimal structural proportion for the demonstrations, proving that MO-IKE is not only effective at preventing structural collapse but also highly sample-efficient to train.

\section{Hyperparameter Sensitivity.} 
\label{appendix: hyperparam}
We conduct additional experiments to check hyperparameter sensitivity for knowledge editing reward weight $\lambda_{ES}$ and constraint weights $\lambda_{PC}, \lambda_{RR}$. We swept $\lambda_{PC}$ and $\lambda_{RR}$ by $100\times$ each, holding the other fixed. Shown by Table~\ref{tab:hyperparameter_sweep}, ESR varies by $\leq 2.0$ points, PC by $\leq 3.3$, RR is unchanged. MO-IKE still outperforms all baselines at every setting, with no monotonic degradation in any metric. The directional pattern matches the design: upweighting PC modestly improves PC (+1.0), while upweighting RR modestly degrades PC (-2.3) without further improving RR, indicating (1, 1) sits in a flat region balancing both objectives.

\setlength{\tabcolsep}{6pt}
\begin{table}[ht]
  \centering
    \resizebox{\columnwidth}{!}{%
  \begin{tabular}{|l|c|c|c|c|}
    \hline
    \textbf{Configuration} & \textbf{S ↑} & \textbf{ES ↑} & \textbf{PC ↑} & \textbf{RR ↑} \\ 
    \hline
    $\lambda_{PC}=100, \lambda_{RR}=1$ & \underline{73.2} & 90.0 & \textbf{80.0} & \textbf{57.7} \\ 
    \hline
    $\lambda_{PC}=1, \lambda_{RR}=100$ & 72.7 & \underline{91.7} & 76.7 & \textbf{57.7} \\ 
    \hline
    $\lambda_{PC}=1, \lambda_{RR}=1$ & \textbf{73.4} & \textbf{92.0} & \underline{79.0} & \textbf{57.7} \\ 
    \hline
  \end{tabular}
  }
  \caption{Hyperparameter sensitivity sweep over the Lagrangian penalty weights.}
  \label{tab:hyperparameter_sweep}
\end{table}

\section{Additional Experiments}
\label{appendix: add_experiments}
\noindent\textbf{Evaluation on Large Datasets.} We expand experiments with the \textsc{UniEdit} \cite{chen2025unieditunifiedknowledgeediting} dataset, a larger dataset composed of 311K editing examples drawn from a diverse mixture of relation types and knowledge domains,
which is roughly a 15$\times$ increase in scale over \textsc{CounterFact}. This setting provides a stricter test of whether MO-IKE's gains generalize beyond the testing of a small benchmark. As indicated by Table~\ref{tab:uniedit}, MO-IKE achieves the best Score and best RR. FactPrompt's high ESR (81.7\%) is misleading: its Score (15.0\%), PC (22.7\%), and RR (7\%) are the lowest in the table, indicating FactPrompt does not let the model effectively learn factual updates — the model parrots the injected editing prompt rather than internalizing new knowledge.

\setlength{\tabcolsep}{6pt}
\begin{table}[ht]
  \centering
  \begin{tabular}{|l|c|c|c|c|}
    \hline
    \textbf{Method}      & \textbf{S ↑} & \textbf{ES ↑} & \textbf{PC ↑} & \textbf{RR ↑} \\ 
    \hline
    FactPrompt           & 15.0 & \textbf{81.7} & 22.7 & 7.0 \\ 
    \hline
    EditCoT              & \underline{21.5} & 43.0 & \textbf{36.0} & \underline{11.0} \\ 
    \hline
    IKE                  & 19.6 & 39.3 & 27.6 & \underline{11.0} \\ 
    \hline
    DR-IKE               & 17.2 & 38.0 & 27.3 & 9.0 \\ 
    \hline
    MO-IKE               & \textbf{24.5} & \underline{51.3} & \underline{28.3} & \textbf{14.3} \\ 
    \hline
  \end{tabular}
  \caption{Evaluation on the UniEdit dataset using Llama-3.2-3B-Instruct. }
  \label{tab:uniedit}
\end{table}

\noindent\textbf{LLM Performance.}Table~\ref{tab:llm_performance} presents the performance of MO-IKE across various frozen LLMs. We observe a general positive scaling trend, where performance of Score (S) improves as model size increases. The highest overall rates are achieved by models with larger parameter scales ($\ge$ 7B). 

\begin{table}[ht]
  \centering
  \resizebox{\columnwidth}{!}{%
  \begin{tabular}{|l|c|c|c|c|}
    \hline
    \textbf{Model (Parameters)}      & \textbf{S↑} & \textbf{ES↑} & \textbf{PC↑} & \textbf{RR↑} \\ 
    \hline
    Qwen 2.5 (7B)                    &75.0 &\textbf{96.0} &88.3  &60.0 \\ 
    \hline
    Qwen 2.5 (1.5B)                  &63.7 &69.7 &67.7 &54.7 \\ 
    \hline
    Llama 3.1 (8B)                   &76.7 &88.0 &77.7 &\textbf{67.3} \\ 
    \hline
    Llama 3.2  (3B)                  & 73.5 & 92.0 & 79.0 & 57.7 \\ 
    \hline
    Mistral v0.3 (7B)                &\textbf{77.4} & 80.1 & \textbf{90.7} & 65.0 \\ 
    \hline
  \end{tabular}
  }
  \caption{Editing performance across different LLMs.}
  \label{tab:llm_performance}
\end{table}

\noindent\textbf{Natural Baselines.} We provide natural baselines that showcase the performance of the retriever, as we claim that selection and ordering and demonstrations matter. Precisely, we compare MO-IKE with random shuffling baseline and a duplication baseline where the target edit is repeated multiple times, while keeping UPDATE and RETAIN unchanged. Table\ref{tab:heuristics} shows that simple duplication or shuffling yields moderate performance but does not match the balance of ES/PC/RR achieved by our multi-objective RL framework.

\setlength{\tabcolsep}{6pt}
\begin{table}[ht]
  \centering
    \resizebox{\columnwidth}{!}{%
  \begin{tabular}{|l|c|c|c|}
    \hline
    \textbf{Method}      & \textbf{ES ↑} & \textbf{PC ↑} & \textbf{RR ↑} \\ 
    \hline
    Random Shuffling     & 77.0--91.0 & 70.0--74.0 & 52.0--60.0 \\ 
    \hline
    Duplicate Target 2$\times$ & 79.3 & 72.3 & 51.0 \\ 
    \hline
    Duplicate Target 4$\times$ & 76.7 & 73.7 & 50.7 \\ 
    \hline
    Duplicate Target 8$\times$ & 78.3 & \underline{77.0} & \underline{52.0} \\ 
    \hline
    Duplicate COPY 2$\times$ & 78.3 & 72.7 & \underline{52.0} \\ 
    \hline
    Duplicate COPY 4$\times$ & 79.0 & 73.7 & 49.7 \\ 
    \hline
    Duplicate COPY 8$\times$ & \underline{79.7} & 70.0 & 51.0 \\ 
    \hline
    MO-IKE               & \textbf{96.0} & \textbf{88.3} & \textbf{60.0} \\ 
    \hline
  \end{tabular}
  }
  \caption{Performance of heuristic prompt engineering strategies on \textsc{CounterFact}.}
  \label{tab:heuristics}
\end{table}

\noindent\textbf{Evaluation on 2025-era-SOTA LLMs.} To directly test the long-term applicability of MO-IKE, we conducted an additional evaluation on a recently released 2025-era model, Qwen3-4B-Instruct. Results are summarized in Table~\ref{tab:qwen3}. We observe that base models exhibit high baseline robustness—for instance, standard static IKE achieves an impressive 98.67\% Edit Success Rate (ESR) and 94.67\% Paraphrase Consistency (PC) on Qwen-3. Since ESR and PC are near the ceiling in this model, RR becomes the key metric in differentiating MO-IKE from all other baselines. We can observe that MO-IKE exceeds the single objective baseline DR-IKE by approximately 23\% on RR, confirming its effectiveness in preserving neighboring facts in knowledge editing. Moreover, MO-IKE successfully preserves the high ESR and PC of the newer models. Therefore, in terms of the overall performance, MO-IKE achieves the highest overall score (S) of 70.2\%.

\setlength{\tabcolsep}{6pt}
\begin{table}[ht]
  \centering
  \begin{tabular}{|l|c|c|c|c|}
    \hline
    \textbf{Method}      & \textbf{S ↑} & \textbf{ES ↑} & \textbf{PC ↑} & \textbf{RR ↑} \\ 
    \hline
    FactPrompt           & 50.4 & 91.0 & 75.3 & 31.3 \\ 
    \hline
    EditCoT              & 42.1 & 86.1 & 76.5 & 24.6 \\ 
    \hline
    IKE                  & \underline{65.6} & \underline{98.7} & \underline{94.7} & \underline{46.3} \\ 
    \hline
    DR-IKE               & 49.6 & \textbf{99.3} & \textbf{96.0} & 30.0 \\ 
    \hline
    MO-IKE               & \textbf{70.2} & 98.3 & 94.3 & \textbf{53.0} \\ 
    \hline
  \end{tabular}
  \caption{Editing performance on Qwen3-4B-Instruct. While stronger base models exhibit high baseline robustness, MO-IKE remains the only framework capable of preserving knowledge specificity (RR), achieving the highest overall score. }
  \label{tab:qwen3}
\end{table}

\noindent\textbf{Computation Cost.} We evaluate inference latency Llama-3.2-3B-Instruct. The results are detailed in Table~\ref{tab:latency}.

\setlength{\tabcolsep}{6pt}
\begin{table}[ht]
  \centering
    \resizebox{\columnwidth}{!}{%
  \begin{tabular}{|l|c|c|c|c|}
    \hline
    \textbf{Component}   & \textbf{Mean (ms)} & \textbf{Median (ms)} & \textbf{Std (ms)} & \textbf{p95 (ms)} \\ 
    \hline
    Retrieval            & 812.2 & 808.5 & 36.2 & 871.8 \\ 
    \hline
    Generation           & 112.3 & 112.8 & 45.6 & 181.3 \\ 
    \hline
    End-to-End           & 924.4 & 915.4 & 55.4 & 1027.3 \\ 
    \hline
  \end{tabular}
  }
  \caption{Inference latency per query evaluated on \textsc{CounterFact} using Llama-3.2-3B-Instruct. Benchmarked on a single NVIDIA A100-SXM4-40GB GPU at batch size 1.}
  \label{tab:latency}
\end{table}

\section{Case Study}
%%% idx: 676
%%% [2. 3. 2. 3. 2. 3. 2. 2. 3. 2. 3. 3. 3. 1. 2. 3. 1. 3. 2. 3. 2. 3. 3. 3. 3. 1. 3. 3. 2. 2. 1. 2.]
\label{appendix: case study}

\begin{table*}[t]
\centering
\small
\begin{tabular}{p{0.12\linewidth} | p{0.26\linewidth} | p{0.26\linewidth} | p{0.26\linewidth}}
\toprule
\textbf{Component} & \textbf{IKE} & \textbf{DR-IKE} & \textbf{MO-IKE} \\
\midrule
\textbf{Edit Request} & \multicolumn{3}{l}{\textit{Target Edit:} Ivan Ivanov-Vano spoke the language
\textbf{Russian} ($y_{true}$) $\rightarrow$ \textbf{French} ($y_{new}$)} \\
\midrule
\textbf{Selected} \newline \textbf{Demonstrations} & 
\textbf{[COPY]} Andrey Malakhov spoke the language French. \newline
\textbf{[COPY]} Boris Shaposhnikov spoke the language French. \newline
\dots \newline
\textbf{[UPDATE]} Andrey Malakhov is a native speaker of French. \newline
\textbf{[UPDATE]} Fyodor Pavlovich Reshetnikov, a native French. \newline
\dots \newline
\textbf{[RETAIN]} \textit{Roger Vitrac spoke the language Russian.} \newline
\textbf{[RETAIN]} \textit{Christophe Moreau spoke the language Russian.} \newline
\dots \newline
\vspace{0.1cm}
\textit{(Note: \textbf{Fixed category order.})} & 
\textbf{[COPY]} Andrey Malakhov spoke the language French. \newline
\textbf{[COPY]} Boris Shaposhnikov spoke the language French. \newline
\dots \newline
\textbf{[UPDATE]} Andrey Malakhov is a native speaker of French. \newline
\textbf{[UPDATE]} Fyodor Pavlovich Reshetnikov, a native French. \newline
\dots \newline
\textit{(Note: \textbf{Fixed COPY and UPDATE order.} No RETAIN candidates, as they are filtered due to penalty on non-edit-maximizing signals)} & 
\textbf{[COPY]} Andrey Malakhov spoke the language French. \newline
\textbf{[RETAIN]} \textit{Christophe Moreau spoke the language Russian.} \newline
\textbf{[UPDATE]} Fyodor Pavlovich Reshetnikov, a native French. \newline
\dots \newline
\textbf{[UPDATE]} Andrey Malakhov is a native speaker of French. \newline
\dots \newline
\textbf{[RETAIN]} \textit{Roger Vitrac spoke the language Russian.} \newline
\textbf{[COPY]} Boris Shaposhnikov spoke the language French. \newline

\vspace{0.1cm}
\textit{(Note: \textbf{Unfixed order.} Balanced structural proportion maintained via soft embedding)} \\
\midrule
\textbf{Reliability Metric} \newline \textit{Ivan Ivanov-Vano spoke the language?} & 
\textcolor{red}{Russian.} (\textit{Unsuccessful edit due to heuristic prompt construction}) & 
\textcolor{blue}{French.} (\textit{Success}) & 
\textcolor{blue}{French.} (\textit{Success}) \\
\midrule
\textbf{Specificity Metric} \newline \textit{Vladimir Smirnov, speaker of?} & 
\textcolor{blue}{Russian.} (\textit{Success}) & 
\textcolor{red}{French.} (\textit{Unsuccessful retention due to unbalanced RETAIN demonstrations}) & 
\textcolor{blue}{Russian.} (\textit{Success}) \\
\bottomrule
\end{tabular}
\caption{Comparison of in-context editing frameworks highlighting the structure of the constructed prompt.}
\label{tab:case_study_comprehensive}
\end{table*}

Table \ref{tab:case_study_comprehensive} use the record 676 in \textsc{Counterfact} to illustrate the relationship between the structural composition of retrieved demonstrations and the resulting performance tradeoffs in different in-context knowledge editing frameworks. We use edit success to represent Reliability Metric, and retention rate to represent Specificity Metric.

\noindent\textbf{\textbf{IKE.}} The baseline IKE framework relies on fixed prompt construction---statically ordering COPY, UPDATE, and RETAIN demonstrations. As shown in the Reliability Metric, this static template approach fails to successfully execute the target edit ($y_{true} \rightarrow y_{new}$). Because the retriever does not optimize for the edit signal, the model defaults to its pre-trained knowledge (Russian) rather than adopting the new fact.

\noindent\textbf{\textbf{DR-IKE.}} 
While DR-IKE successfully forces the edit (outputting ``French'' for Ivanov-Vano), it suffers from a ``over-optimization'' issue. By penalizing only on non-edit-maximizing signals, the policy optimization entirely discards RETAIN candidates. This creates an unbalanced prompt heavily skewed toward the target concept. Consequently, DR-IKE exhibits fact forgetting, failing the Specificity metric by incorrectly altering the language of a related but distinct entity (Vladimir Smirnov) to French.

\noindent\textbf{\textbf{MO-IKE.}} MO-IKE resolves the tradeoff between edit success and knowledge retention. By avoiding rigid templates and maintaining a balanced structural proportion of demonstration types, MO-IKE allows the retriever to dynamically interleave UPDATE, RETAIN, and COPY examples. This unfixed ordering provides sufficient contextual protection for unrelated facts while still providing a strong enough edit signal to update the target knowledge. As a result, MO-IKE is the only framework in the case study that successfully achieves both Reliability (updating Ivanov-Vano to French) and Specificity (retaining Russian for Smirnov).

\end{document}